\documentclass[final]{cvpr}

\usepackage{times}
\usepackage{epsfig}
\usepackage{graphicx}
\usepackage{amsmath}
\usepackage{amssymb}

\usepackage{caption}
\usepackage{subcaption}

\usepackage[pagebackref=true,breaklinks=true,colorlinks,bookmarks=false]{hyperref}

\def\cvprPaperID{8884} 
\def\confYear{arXiv}
\begin{document}

\title{An Empirical Method to Quantify the Peripheral Performance Degradation in Deep Networks}

\author{Calden Wloka \qquad John K. Tsotsos\\
Department of Electrical Engineering and Computer Science\\
York University, Toronto, Canada\\
{\tt\small calden, tsotsos@eecs.yorku.ca}
}

\maketitle

\begin{abstract}

  When applying a convolutional kernel to an image, if the output is to remain the same size as the input then some form of padding is required around the image boundary. This means that for each layer of convolution in a convolutional neural network (CNN), a strip of pixels equal to the half-width of the kernel size is produced with a non-veridical representation. Although most CNN kernels are kept small to reduce the parameter load of a network, this non-veridical area compounds with each convolutional layer. The tendency toward deeper and deeper networks combined with stride-based down-sampling means that the propagation of this region through the network can end up covering a non-negligable portion of the image. Although this issue with convolutions has been well acknowledged over the years, the impact of this degraded peripheral representation on modern network behavior has not been fully quantified. What are the limits of translation invariance? Does image padding successfully mitigate the issue, or is performance affected as an object moves between the image border and center? Using Mask R-CNN as an experimental model, we design a dataset and methodology to quantify the spatial dependency of network performance. Our dataset is constructed by inserting objects into high resolution backgrounds, thereby allowing us to crop sub-images which place target objects at specific locations relative to the image border. By probing the behaviour of Mask R-CNN across a selection of target locations, we see clear patterns of performance degredation near the image boundary, and in particular in the image corners. Quantifying both the extent and magnitude of this spatial anisotropy in network performance is important for the deployment of deep networks into unconstrained and realistic environments in which the location of objects or regions of interest are not guaranteed to be well localized within a given image.

\end{abstract}

\section{Introduction}

Convolution operations have long been a mainstay of image processing and computer vision, and even as deep learning techniques grow more complex and varied in design convolution remains at the heart of the majority of deep networks designed for visual applications. Convolutions have a number of attractive properties for deep learning. Convolution layers require drastically fewer parameters than fully connected layers, making them both more memory efficient and less prone to overfitting. Additionally, the spatial constraints of a kernel restrict feature generation to a local neighbourhood (analogous to the receptive field of a biological neuron). Tsotsos \cite{Tsotsos1987} first argued that hierarchical layers of convolutions were not only well-suited to visual processes by ameliorating the combinatorics of vision, but also matched the basic architecture of biological visual cortex. Ulyanov \etal \cite{UlyanovEtAl2018} recently revived this argument in the context of modern deep networks.

\begin{figure}[!htbp]
	\begin{center}
		\includegraphics[width=0.9\linewidth]{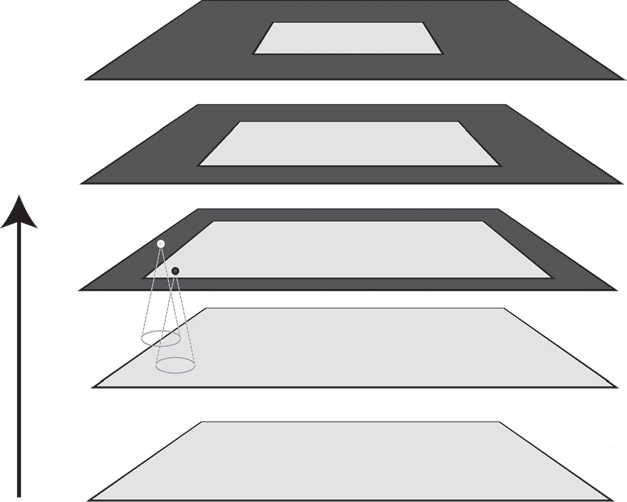}
	\end{center}
	\caption{Diagram of the growth in the region of an input image (bottom layer) with undefined convolution results (shown in dark grey) over successive applications of convolutions. Figure adapted from \cite{Tsotsos2011Book}.
	 \label{fig:BoundaryProblem}}
\end{figure}

Another highly attractive property of convolution is that it is spatially invariant to translation, allowing the same visual pattern to be detected regardless of its location within a visual scene. There are some important caveats to this point, however. Zhang \cite{Zhang2019} showed that max pooling and stride techniques, which are widely used in deep learning to reduce the computational load of a network, can create local performance anisotropies. More fundamental to convolution itself, when a kernel is centered on a pixel near the border of the image, part of the kernel will extend beyond the bounds of the image and the results of the convolution will be undefined. The most straightforward approach to this issue is to shrink the size of the output to only include pixels with defined convolution values. This discards a number of pixels equal to the kernel half-width along each image border. For a single convolution with a small kernel relative to the image size, this reduction can often be negligible. However, when applying convolution kernels to very small images (\eg the top layers of an image pyramid \cite{vanderWalBurt1992}), the loss of border pixels may be a substantial portion of one's input. A similar problem arises when stacking many convolutions over the same input, as in very deep networks (\eg \cite{HeEtAl2016}). Even if each kernel is very small, the cumulative loss of pixels over many layers could likewise amount to a prohibitive loss (see Figure \ref{fig:BoundaryProblem}).

An alternative to shrinking the output is to pad the input to a convolutional layer in order to maintain an output size equal to that of the input. A number of common strategies are available, such as zero-padding (the input image is extended with zeros) and replicate (extended pixels copy the value of the closest pixel from the original input). Although padding allows the size of the output layer to equal the size of the input, the output of convolutions conducted near the image border will clearly be affected by the values supplied by the padding, thereby potentially degrading performance within this region. Multiple recent works \cite{IslamEtAl2020,KayhanGemert2020} have demonstrated measurable effects on the internal representations of a convolutional network due to the presence of padding, noting that pixel padding provides a signal that allows a network to encode absolute spatial position in an image. However, to our knowledge no work has examined the effect that image padding has on the translation invariance of deep networks. Given the importance of translation invariance to network generalization in unconstrained environments, it is critical for many applications to verify this assumption and quantify any limits which exist.

\subsection{Motivation}
\label{subsec:Motivation}

\begin{figure}[!htbp]
	\begin{center}
		\includegraphics[width=0.9\linewidth]{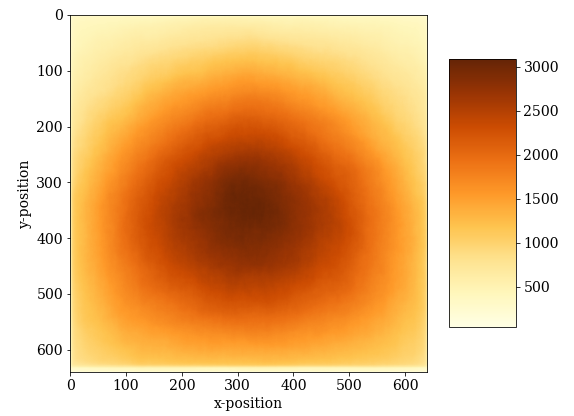}
	\end{center}
	\caption{A plot of the sum of binary object masks from the MS-COCO 2017 validation set after resizing all images in the set to $640 \times 640$ pixels. Darker values correspond to a greater density of annotated objects overlapping that location. It is clear that objects are most frequently located just below the image center, and rarely appear along the image border (particularly the upper corners).
	 \label{fig:coco_dist}}
\end{figure}

Standard benchmark datasets often have strong structural biases toward well-framed image compositions which place objects of interest away from the image boundary and toward the center of the image, commonly known as \emph{photographer bias} \cite{TsengEtAl2009}. Figure \ref{fig:coco_dist} shows the density of object masks in the Microsoft Common Objects in Context (MS-COCO) 2017 validation set \cite{LinEtAl2014}, constructed by resizing all images to a standardized shape ($640 \times 640$) and then summing the binary masks of all annotated objects onto a single map. As can be seen, the vast majority of annotated objects are located in the lower central portion of the image, and the image borders (and corners in particular) contain very few target objects. MS-COCO has remained a standard dataset for performance benchmarking and is regularly used in annual competitions to determine state of the art in object detection and segmentation. Given the distribution of target objects within the dataset, however, network performance over objects near an image periphery cannot be easily predicted and will minimally factor into comparative rankings between competing network designs. Nevertheless, when applying deep networks to unconstrained environments, it is very important to know about any spatial anisotropies in performance so that these may be factored into the design.

For example, due to the elevated angle of image capture as well as potential shifts caused by wind, aerial based image acquisition, such as traffic monitoring \cite{MariaEtAl2016}, has a greatly increased chance of capturing images with targets of interest across the entire field of view of a camera. Knowing the spatial pattern of performance can allow for more optimal control strategies to acquire subsequent frames and reduce the chance of missed targets. Similarly, for remote sensing applications such as crop monitoring \cite{BallesterosEtAl2018}, regions of interest can show up in any image location, and it therefore becomes useful to quantify the spatial properties of a network when designing the data acquisition strategy.

Spatial performance quantification can similarly be important for optimal design in more constrained setups. For example, when using a camera array to provide coverage of specified area (such as the surroundings of a vehicle \cite{Corcoran2017}), it would be useful to incorporate network behavior when determining the overlap of the fields of view of neighboring cameras, thereby ensuring performance can be maintained evenly without unexpected localized drops while also minimizing redundant processing. Likewise, when specific areas of the environment are known to be important for accomplishing a task, quantification of spatial anisotropy can allow a system to be designed to ensure that the acquired field of view keeps high priority areas within the optimal performance region of an image. For example, cameras mounted on shopping carts have been proposed for detecting out of stock products \cite{ChaubardGarafulic2018}, but if the top or bottom shelves frequently fall only within the periphery of the camera's field of view there may be lower detection rates for these shelves. Similarly, for safety monitoring using fixed camera setups (\eg person overboard detection \cite{KatsamenisEtAl2020}), it can be vital to ensure that the fields of view of the system cameras well-localize critical regions of interest.

A final application in which spatial anisotropy of network performance is important is in domains relying on high resolution imagery beyond the memory capacity of the available computing power. This can be due to limited onboard processing (such as on autonomous drones \cite{PlastirasEtAl2018}) or due to ultra high resolution input images (such as is commonly found in remote sensing applications \cite{ChenEtAl2019}). Tiled subcrops of a high resolution image can be used to reduce the memory load of a single network pass to a more manageable level, but if the tiles do not sufficiently overlap this may cause a drop in network performance along the grid lines of the tiling.

\subsection{Contribution}

In this paper we verify the existence of spatial anisotropies in deep convolutional network performance, empirically demonstrating limitations on the assumption of translation invariance. We do this by designing a novel dataset and testing methodology using composite images in order to quantify the extent and magnitude of spatial anisotropy in deep neural networks using parametrically controlled presentations of single targets inserted into high resolution backgrounds (Section \ref{sec:Methods}).

We demonstrate this methodology in the semantic segmentation domain, characterizing the behavior of the Mask R-CNN network \cite{HeEtAl2017} and showing peripheral deficits in target prediction rates and network confidence scores, particularly for smaller targets (Section \ref{sec:Results}). Although we concentrate on a specific example network and application domain, the anisotropic behavior characterized by our methodology is unlikely unique to Mask R-CNN or semantic segmentation, and this empirical methodology to quantifying spatial anisotropy in network performance can be rather straightforwardly adapted to other problem domains. This will be a critical step to system design and optimization for many real-world applications.

\section{Methods}
\label{sec:Methods}

The primary goal of our method is to quantify translation invariance in network performance. The first step to accomplish this goal is to construct a dataset which allows one to parametrically control the spatial arrangement and size of target objects presented to the network in order to probe the behavior of the network with respect to the size and location of objects within a scene. Although previously Rosenfeld \etal examined shifts in network behavior in response to object translation \cite{RosenfeldEtAl2018}, in that study the authors translated an object within a scene and observed changes in network behavior hypothesized to be a result of crosstalk between object representations within the network. In order to focus explicitly on the effect of spatial location on network behavior, we instead keep a given target object static with respect to the scene content and shift the field of view presented to the network in order to place the object at the desired location. All scenes are also composed with only one object corresponding to the set of target classes the network has been trained to recognize, thereby reducing any confounding representational interactions like those demonstrated by Rosenfeld \etal \cite{RosenfeldEtAl2018} and predicted in \cite{Tsotsos2011Book}.

\subsection{Network}

For this experiment we selected the Mask R-CNN network \cite{HeEtAl2017}. Mask R-CNN is a widely used network for object detection and segmentation, and still forms a foundation for current state of the art benchmark leaders \cite{LiEtAl2020}. Using our methodology to characterize and quantify spatial anisotropy in Mask R-CNN performance therefore demonstrates that this is an issue still relevant to modern architectures even when they are not formulated as a classic fully convolutional neural network.

We deliberately chose to use the default pre-trained implementation of Mask R-CNN provided by the \texttt{torchvision} package\footnote{Version 0.5.0}. This makes our results directly applicable to the widest audience, and ensures that any anisotropic behavior we find is not due to a property we introduced.

\subsection{Dataset Construction}
\label{subsec:Dataset}

\begin{figure*}[!htbp]
	\begin{center}
    \begin{subfigure}[b]{0.3\textwidth}
      \includegraphics[width=\textwidth]{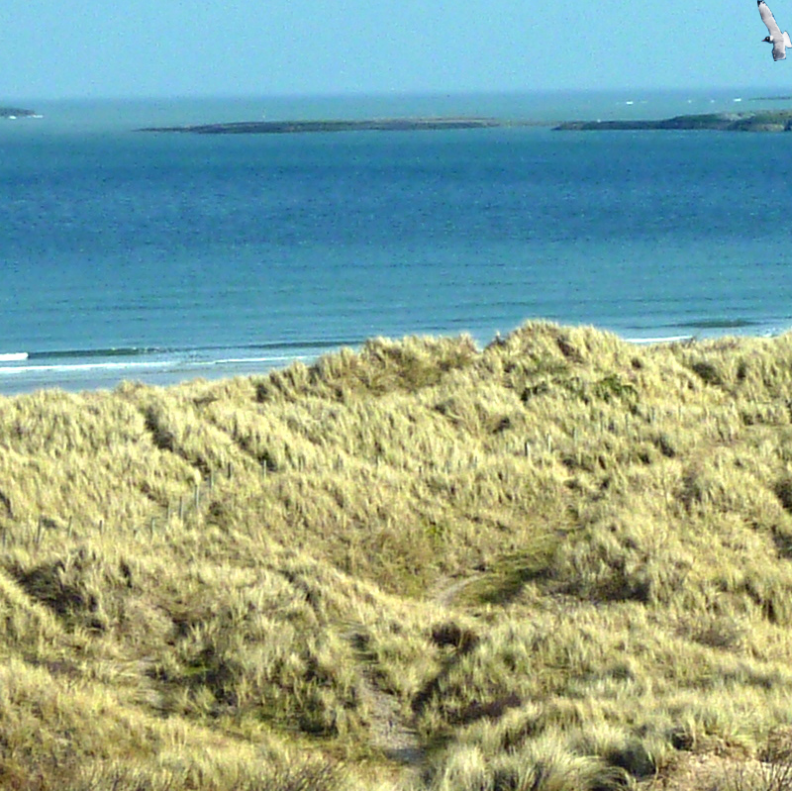}
      \caption{A bird target proportional to $0.08$ the crop dimensions with a $(0,0)$ pixel offset from the image boundary}
      \label{subfig:bird_corner}
    \end{subfigure}
    \hfill
    \begin{subfigure}[b]{0.3\textwidth}
      \includegraphics[width=\textwidth]{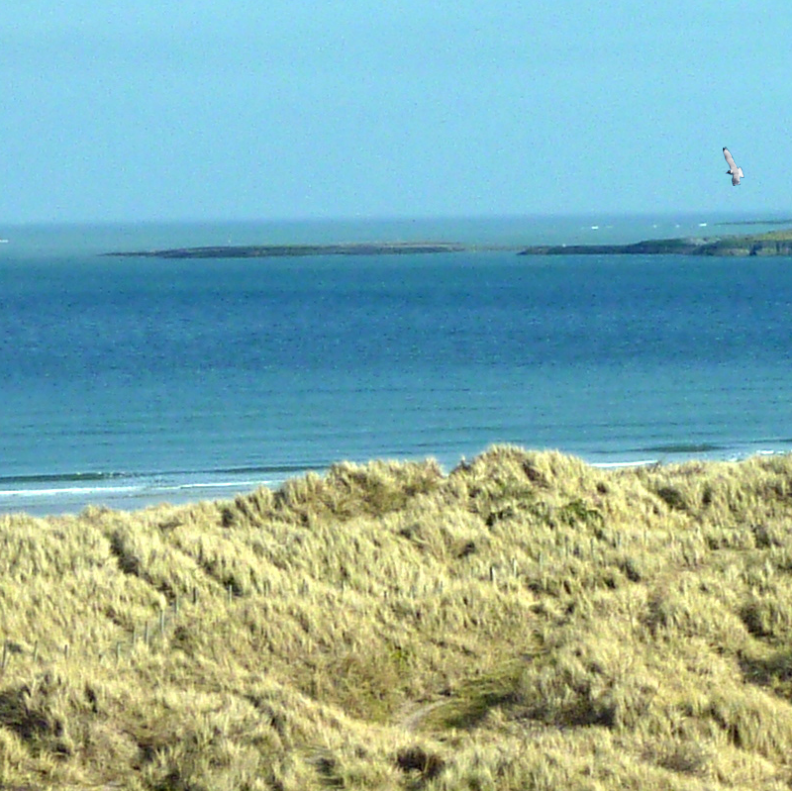}
      \caption{A bird target proportional to $0.05$ the crop dimensions with a $(50,150)$ pixel offset from the image boundary}
      \label{subfig:small_bird}
    \end{subfigure}
    \hfill
    \begin{subfigure}[b]{0.3\textwidth}
      \includegraphics[width=\textwidth]{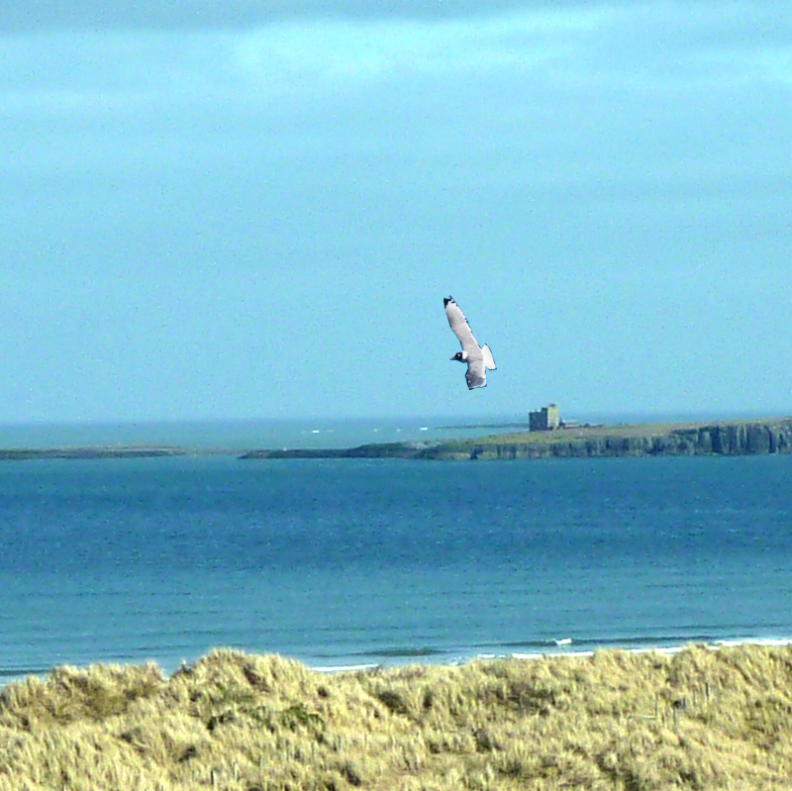}
      \caption{A bird target proportional to $0.12$ the crop dimensions with a $(300,300)$ pixel offset from the image boundary}
    \end{subfigure}
  	\vspace{-4mm}
	\end{center}
	\caption{An example set of test images generated by our dataset showing a bird target inserted into a coastal background at three different positions and sizes. Pixel offsets between the bird and the image boundary are relative to the upper right corner. Note that the bird's position within the scene remains constant, and the field of view of the cropped image shifts to place the target closer or further from the image boundary.
	 \label{fig:ImgExamples}}
\end{figure*}

Our dataset consists of the following components: \emph{background images} and \emph{target objects}. Each test image in the dataset is generated by resizing a target object according to a specified dimension and inserting that object into a randomly selected ``insertion location'' from a set of contextually valid candidate locations in a background image (\ie, locations where the target object might commonly be found). Insertion locations are sampled in the coordinate frame of the full background image, and remain constant for a given target object once it has been generated. When evaluating network behavior for a given a test image, ``probe location'' determines where in the sub-image cropped from the test image the target object will be, as measured by the distance between the target object and the sub-image boundaries. Figure \ref{fig:ImgExamples} gives an example set of cropped images sampled from a test image based on a composite of a flying bird target object and a coastal background. Note that the insertion location remains constant (the bird's position within the larger background is the same in each crop), whereas the probe position and target size vary.

Further design details governing background images and target objects are given in Sections \ref{subsec:Backgrounds} and \ref{subsec:Targets}, respectively, and the network examination protocol is given in Section \ref{subsec:Experiment}. It should be noted that while our design strongly attempts to avoid visual artifacts and generate images which look plausible to a human on first glance, we are not enforcing strict compositional realism. Our use of composite test images generated by combining natural image components from multiple sources was governed primarily by the simplicity and low computational requirements of such an approach, as opposed to generating fully synthetic test images (\eg see \cite{Wrenninge2018,martinezgonzalez2019} for examples of work in this area) which could potentially achieve a similar level of parametric control while also solving some of the issues of inconsistent lighting, viewpoint, and spatial scale caused by directly combining elements from independent source images.

\subsubsection{Background Images}
\label{subsec:Backgrounds}

$34$ high resolution natural scenes licensed for research purposes or free use were chosen to serve as the background set for this dataset. Each image has a minimum size of $1600$ pixels in both height and width; this is to ensure that it will always be possible to take a $800 \times 800$ pixel crop of the image which places the target object in one of the corners of the cropped image, regardless of where in the image that object is placed. This crop size is necessary to supply images of the expected input resolution to Mask R-CNN. When determining the target object size relative to the cropped image, we use \emph{crop dimension} to refer to the major image dimension. Since the cropped input in this case is square, this is equal to $800$ pixels.

Background images were visually inspected to ensure no existing scene elements corresponded to the list of recognizable target categories in the MS-COCO dataset\footnote{person, bicycle, car, motorcycle, airplane, bus, train, truck, boat, traffic light, fire hydrant, stop sign, parking meter, bench, bird, cat, dog, horse, sheep, cow, elephant, bear, zebra, giraffe, backpack, umbrella, handbag, tie, suitcase, frisbee, skis, snowboard, sports ball, kite, baseball bat, baseball glove, skateboard, surfboard, tennis racket, bottle, wine glass, cup, fork, knife, spoon, bowl, banana, apple, sandwich, orange, broccoli, carrot, hot dog, pizza, donut, cake, chair, couch, potted plant, bed, dining table, toilet, tv, laptop, mouse, remote, keyboard, cell phone, microwave, oven, toaster, sink, refrigerator, book, clock, vase, scissors, teddy bear, hair drier, toothbrush}. Each image was annotated to include ``insertion regions'', each of which has an associated subset of MS-COCO objects which are consistent with the environment contained in the region (for example, in an insertion region placed over a region of sky, an airplane would be conceptually plausible but a bear would not). In order to create more conceptually consistent images, a small number of MS-COCO categories were broken into sub-categories for the purposes of specifying valid insertion locations: the MS-COCO ``bird'' category was split into \emph{bird\_walking}, \emph{bird\_flying}, and \emph{bird\_swimming}, and the MS-COCO ``boat'' category was split into \emph{boat} and \emph{ship}. When evaluating performance, these categories were collapsed back together to be consistent with the trained categories of the network. One additional constraint of note on the structure of insertion regions is the need to avoid placing objects in locations which might lead to obvious visual artifacts like straddling occluding scene elements; therefore multiple disjoint regions in the same image might encode the same object category in order to avoid environmental features which would cause obvious visual artifacts.

\begin{figure*}[!htbp]
	\begin{center}
    \begin{subfigure}[b]{0.3\textwidth}
      \includegraphics[width=\textwidth]{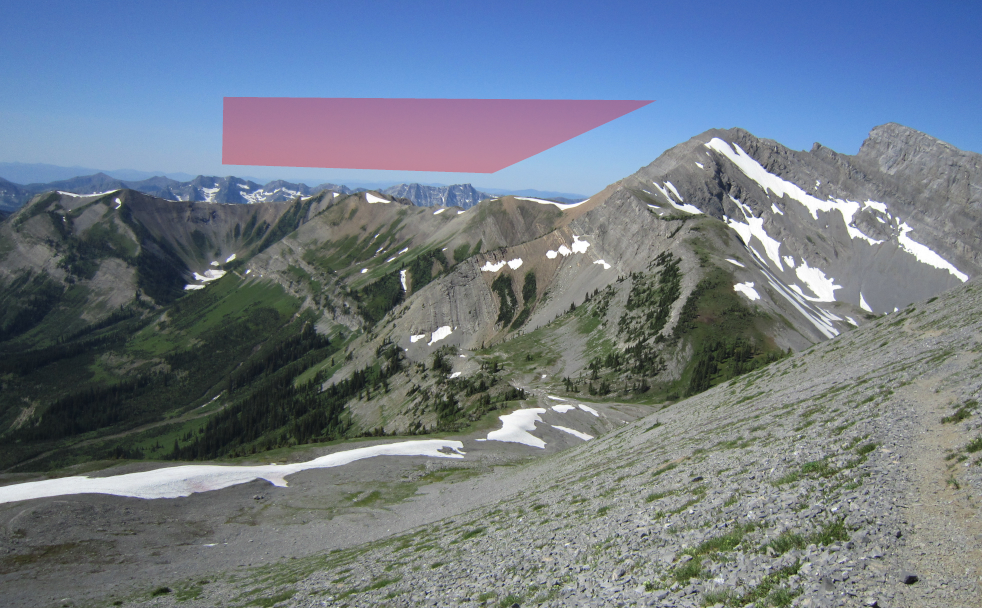}
      \caption{Insertion region for objects from the \emph{bird\_flying} and \emph{airplane} classes}
    \end{subfigure}
    \hfill
    \begin{subfigure}[b]{0.3\textwidth}
      \includegraphics[width=\textwidth]{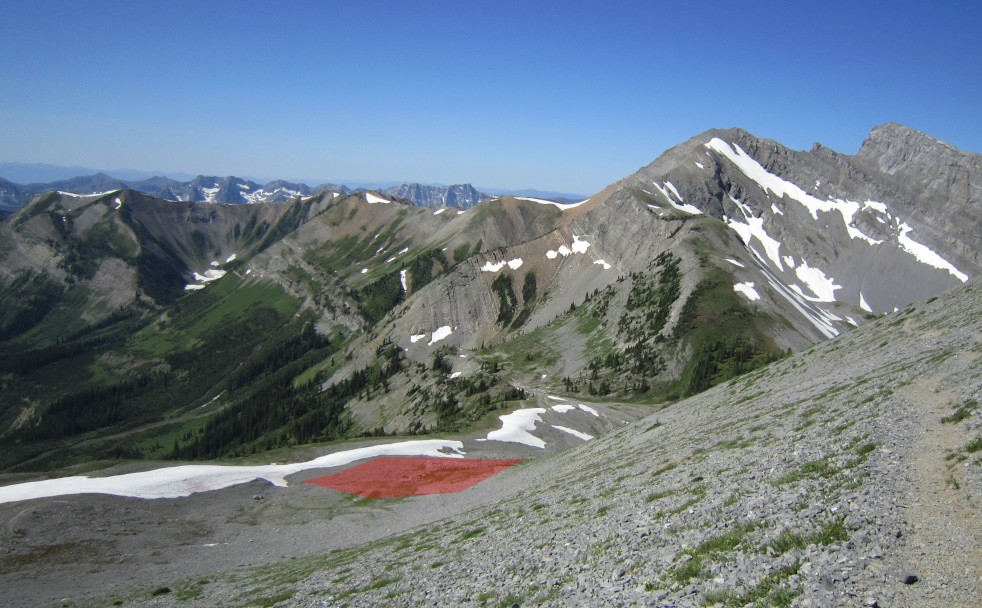}
      \caption{Insertion region for objects from the \emph{cow} and \emph{bear} classes}
    \end{subfigure}
    \hfill
    \begin{subfigure}[b]{0.3\textwidth}
      \includegraphics[width=\textwidth]{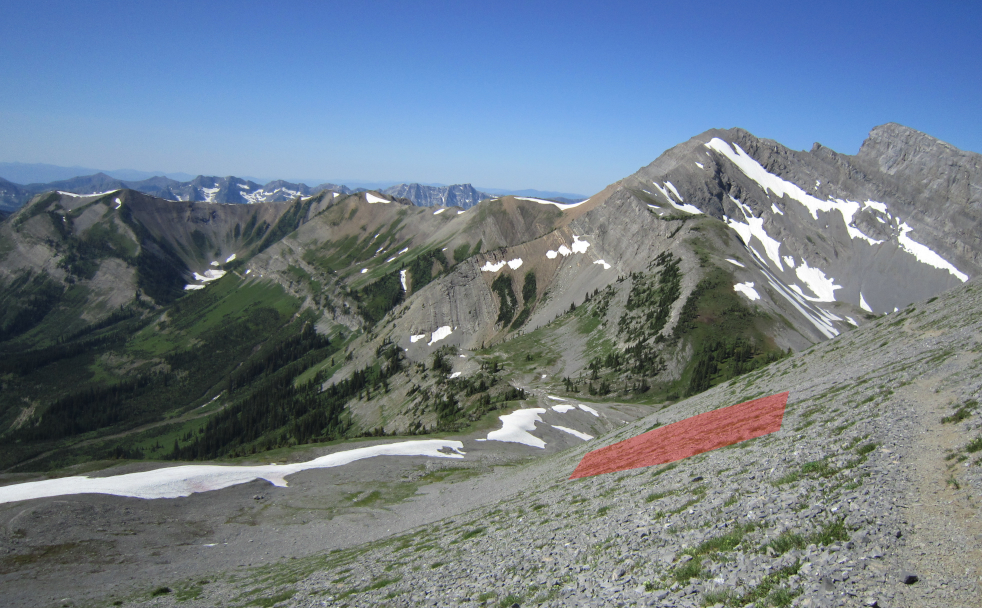}
      \caption{Insertion region for objects from the \emph{sheep}, \emph{cat}, \emph{bird\_walking}, and \emph{dog} classes}
    \end{subfigure}
  	\vspace{-4mm}
	\end{center}
	\caption{An example of the three insertion regions specified for a mountain background. In each image, a different insertion region is highlighted in red.
	 \label{fig:SpawnRegions}}
\end{figure*}

Although insertion regions are not annotated with a minimal distance from the image boundary, they are concentrated as much as possible toward the central portions of the image to better support the necessary crops. When generating target images, all portions of insertion regions within $400$ pixels of the image boundary are suppressed in order to ensure objects may be cropped sufficiently well centered for all probe positions in the experiment. This limit was not made a fundamental aspect of insertion region annotation, however, in order to allow this dataset to more easily be used to quantify spatial anisotropy in additional networks.

\subsubsection{Target Objects}
\label{subsec:Targets}

\begin{figure*}[!htbp]
	\begin{center}
    \begin{subfigure}[t]{0.13\textwidth}
      \includegraphics[width=\textwidth]{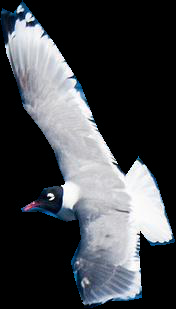}
      \caption{Bird\_flying}
    \end{subfigure}
    \hfill
    \begin{subfigure}[t]{0.2\textwidth}
      \includegraphics[width=\textwidth]{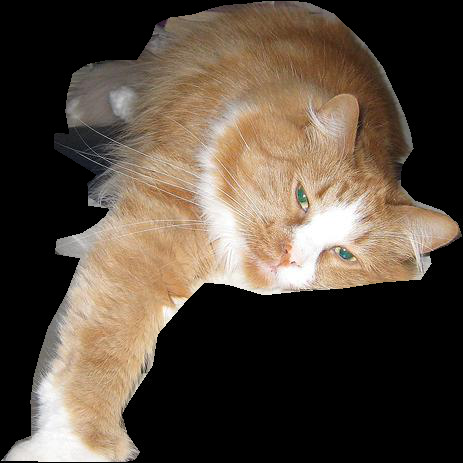}
      \caption{Cat}
    \end{subfigure}
    \hfill
    \begin{subfigure}[t]{0.14\textwidth}
      \includegraphics[width=\textwidth]{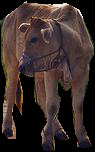}
      \caption{Cow}
    \end{subfigure}
    \hfill
    \begin{subfigure}[t]{0.14\textwidth}
      \includegraphics[width=\textwidth]{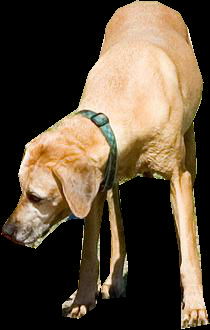}
      \caption{Dog}
    \end{subfigure}
    \hfill
    \begin{subfigure}[t]{0.2\textwidth}
      \includegraphics[width=\textwidth]{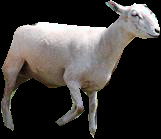}
      \caption{Sheep}
    \end{subfigure}
  	\vspace{-4mm}
	\end{center}
	\caption{A random set of example target objects selected for the dataset. All pixels not contained within the object mask have been set to black.
	 \label{fig:TargetObjects}}
\end{figure*}

Target objects are selected from $14$ of the available MS-COCO classes. Target classes were selected in order to provide a mixture of object types (such as animal, vehicle, and small object) which would be conceptually consistent with the background images and for which well segmented exemplars could be found in the MS-COCO dataset. Candidate objects were first collected based on their native resolution (objects with largest dimension below $50$ pixels were rejected as too likely to introduce artifacts when resized and inserted into background images), and then manually inspected. Any candidate object mask which contained a large quantity of extraneous content from the original MS-COCO image was rejected (the entire ``bicycle'' class, for example, was rejected due to the widespread retention of background textures through the spokes and bodies of the bicycles). Likewise, if a large portion of the candidate object was missing due to occlusion by another scene element in the original image, it was rejected. The goal was to create a subset of target objects which could be recognizably inserted with minimal visual artifacts into the the insertion locations annotated on the background images. As mentioned in Section \ref{subsec:Backgrounds}, the MS-COCO ``bird'' category was split into \emph{bird\_walking}, \emph{bird\_flying}, and \emph{bird\_swimming}, and the MS-COCO ``boat'' category was split into \emph{boat} and \emph{ship}. See Figure \ref{fig:TargetObjects} for some examples of target objects accepted into the dataset.

In total, $284$ candidate objects were collected in the following categories (with the number of target objects given in parentheses): \emph{airplane} (12), \emph{bear} (18), \emph{bench} (10), \emph{bird\_flying} (3), \emph{bird\_walking} (18), \emph{bird\_swimming} (1), \emph{boat} (7), \emph{ship} (9), \emph{cat} (52), \emph{car} (21), \emph{cow} (27), \emph{dog} (30), \emph{frisbee} (5), \emph{handbag} (19), \emph{sheep} (26), \emph{skateboard} (13), and \emph{suitcase} (13).

\subsection{Experiment Setup}
\label{subsec:Experiment}

For each target object with a valid insertion location in a given background image, we generated a random location from that insertion region, resulting in $3980$ unique background and target pairs. For each of these pairs, objects were inserted after resizing based on the ratio of the largest dimension of the target object to the width of the image crop ($800$ pixels). Target objects were inserted at the following sizes relative to the cropped image: $0.05, 0.08, 0.12, 0.18$, which corresponds to a major dimension of $40, 64, 96, 144$ pixels, respectively, and $15920$ total test images.

Image crops were generated over each test image based on the $(x,y)$ position of the target object relative to the image border. These values were always computed relative to the closest corner of the full background image, and thus a positive $x$ position could correspond to a leftward or rightward shift of the target object (and likewise for the $y$ position value) depending on the global layout of the test image. No significant difference was found in network behaviour with respect to whether the object was being positioned relative to the left or right image border in the $x$ direction, and likewise whether it was relative to the top or bottom image border in the $y$ direction, so the results have been collapsed to simply an absolute $(x,y)$ offset from the cropped image border.

Due to the prohibitive computational requirements of densely testing every possible target position between the corner and center of the cropped images for all test images, a sparse set of offset positions was probed, with probed positions more densely located near the image edge (where the performance is expected to be least uniform). For the smallest target object sizes, $x$ and $y$ positions were set to 0, 2, 4, 7, 10, 14, 18, 24, 30, 38, 46, 60, 75, 90, 120, 150, 200, 250, 300, and 350 pixels. For target objects with major dimension greater than $50$, the final offset value ($350$) was not probed, since an offset of $350$ would position the target object beyond the center of the image and potentially cause the crop to extend beyond the bounds of the test image. Likewise, for the largest target objects both the offset magnitudes of $300$ and $350$ pixels were skipped.

\subsection{Evaluation}

Network performance was evaluated separately for the top prediction (regardless of the accuracy of the predicted class), $t$, and for the top prediction with an accurate class label, $a$. Predictions which do not overlap the target object at all are discarded prior to scoring. For each offset coordinate pair, we compute the prediction rate ($r_t$ for the top prediction rate, and $r_a$ for the accurate prediction rate) as the proportion of test images with a qualifying prediction. Thus, $r_t$ is simply the proportion of test images for which \emph{any} prediction overlaps the target, whereas $r_a$ is equal to the proportion of test images for which a prediction with an accurate class label overlaps the target object.

Additionally, for each position we compute the average prediction confidence, $c$, and intersection over union (IoU) score, $s$. Note that these values are averaged only over valid predictions so as to evaluate the quality of the predictions which are made rather than confound these measures with the positional miss rate.

\section{Results}
\label{sec:Results}

Results for target objects of size $0.05$, $0.08$, $0.12$, and $0.18$ times the crop dimension are shown in Figures \ref{fig:res_5prop}, \ref{fig:res_8prop}, \ref{fig:res_12prop}, and \ref{fig:res_18prop}, respectively. The discussion of results has been split between small objects (size $0.05$ and $0.08$ times the crop dimension) in Section \ref{subsec:small_objects} and large objects (size $0.12$ and $0.18$ times the crop dimension) in Section \ref{subsec:large_objects}.

\subsection{Results for Small Objects}
\label{subsec:small_objects}

\begin{figure*}[!htbp]
	\begin{center}
    \begin{subfigure}[t]{0.45\textwidth}
      \includegraphics[width=\textwidth]{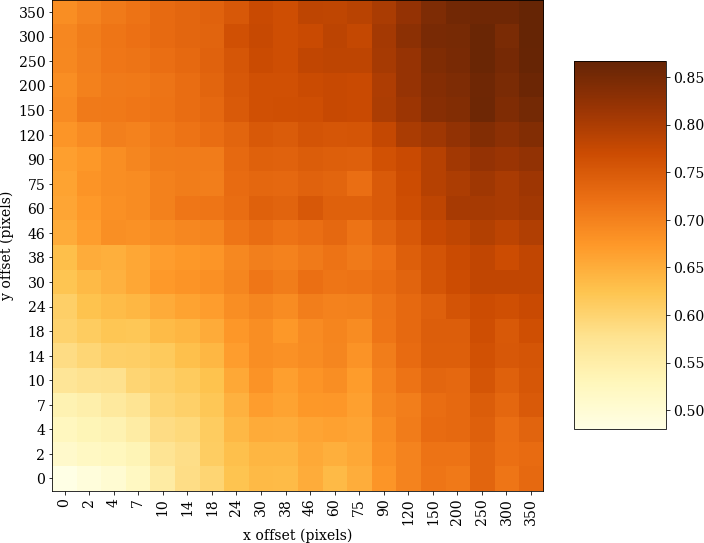}
      \caption{Prediction rate, $r_t$ \label{subfig:rt_05}}
    \end{subfigure}
    \hfill
    \begin{subfigure}[t]{0.45\textwidth}
      \includegraphics[width=\textwidth]{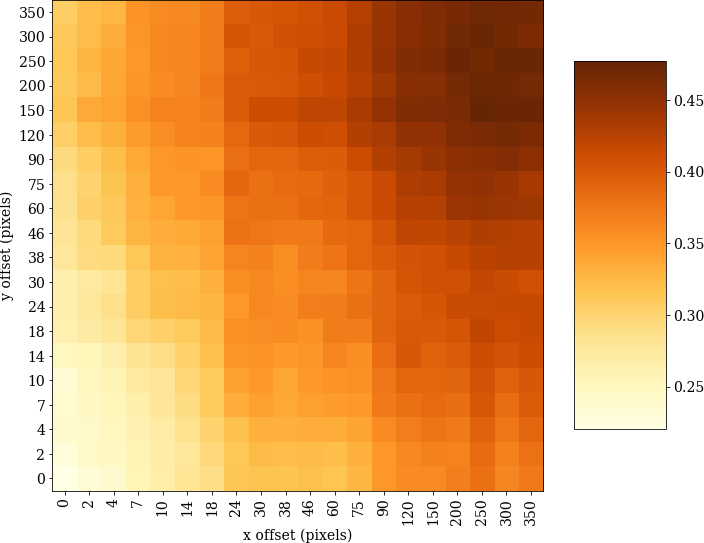}
      \caption{Accurately labeled prediction rate, $r_a$ \label{subfig:ra_05}}
    \end{subfigure}
    \begin{subfigure}[t]{0.45\textwidth}
      \includegraphics[width=\textwidth]{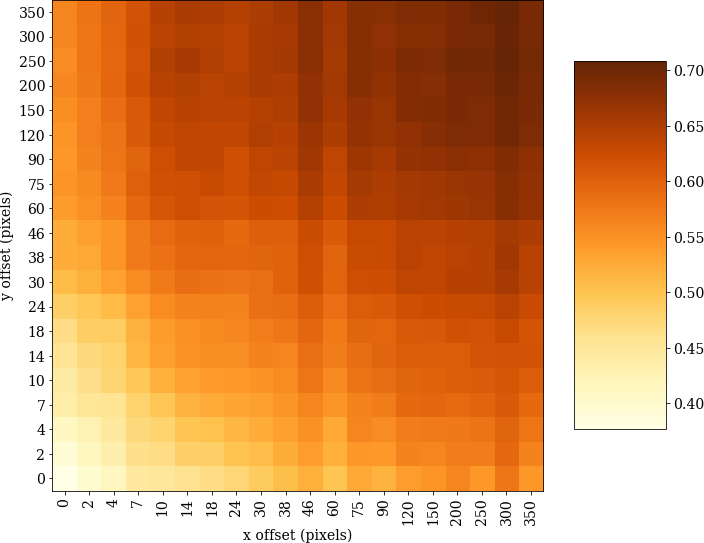}
      \caption{Average confidence score for the top prediction, $c_t$ \label{subfig:ct_05}}
    \end{subfigure}
    \hfill
    \begin{subfigure}[t]{0.45\textwidth}
      \includegraphics[width=\textwidth]{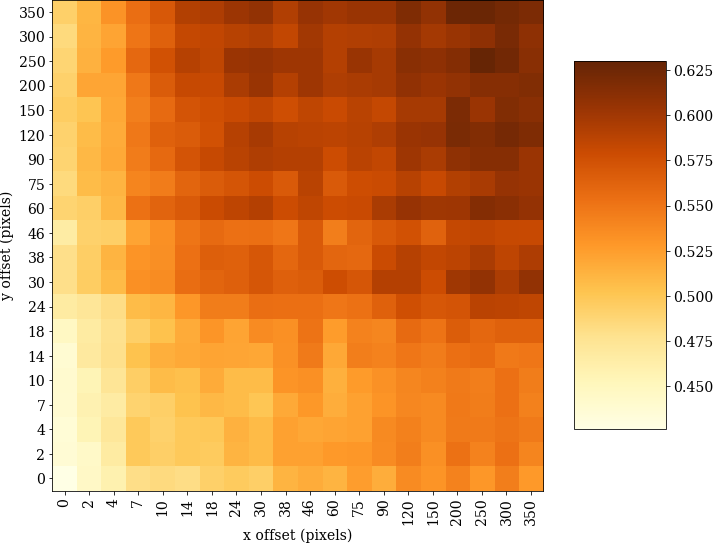}
      \caption{Average confidence score for the top accurately labeled prediction, $c_a$ \label{subfig:ca_05}}
    \end{subfigure}
    \begin{subfigure}[t]{0.45\textwidth}
      \includegraphics[width=\textwidth]{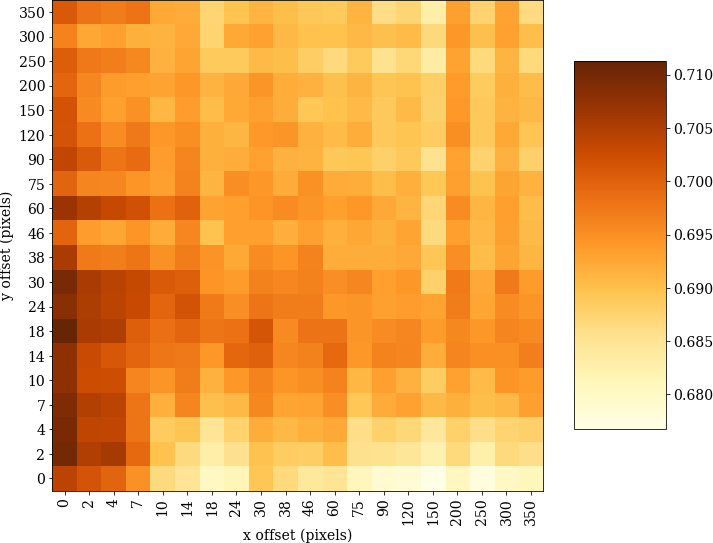}
      \caption{Average IoU score of the top prediction, $s_t$ \label{subfig:st_05}}
    \end{subfigure}
    \hfill
    \begin{subfigure}[t]{0.45\textwidth}
      \includegraphics[width=\textwidth]{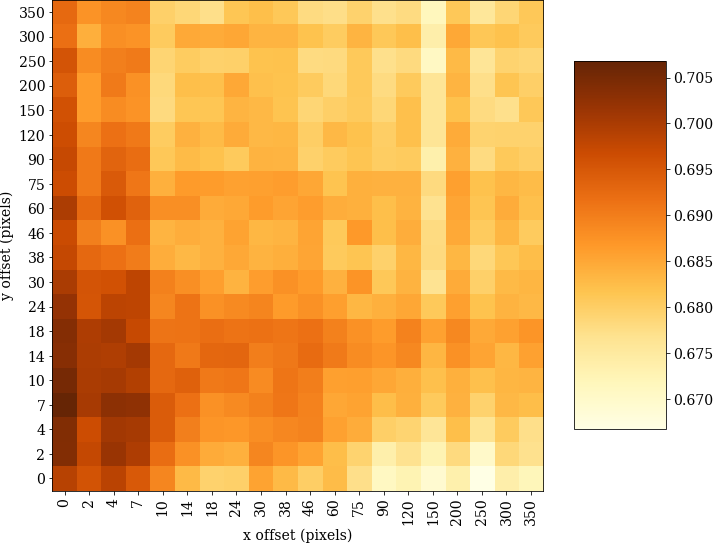}
      \caption{Average IoU score of an accurately labeled prediction, $s_a$ \label{subfig:sa_05}}
    \end{subfigure}
	\end{center}
  \caption{Mask R-CNN performance with respect to spatial position for objects with major dimension equal to $0.05$ times the width of the image crop ($40$ pixels).
	 \label{fig:res_5prop}}
\end{figure*}

Both the rate of valid prediction with any class label and accurately labeled predictions show a consistent and clear pattern of deficit when the target object approaches the image boundary. This deficit is compounded in the image corner and, unsurprisingly, is greater in magnitude for smaller targets, as larger objects extend a greater proportion of pixels outside the region of processing affected by image padding. Interestingly, for the smallest target size, the detection deficit, while greatest near the image boundary, is still measurably present up to $75$ pixels in from the image border as an approximately $10\text{-}15\%$ percentage point drop in valid prediction rate (Figure \ref{subfig:rt_05} and a $5\text{-}10\%$ percentage point drop in accurate prediction rate \ref{subfig:ra_05}. For an $800 \times 800$ pixel image, this still corresponds to a distinct performance deficit over more than a third of the image.

\begin{figure*}[!htbp]
	\begin{center}
    \begin{subfigure}[t]{0.45\textwidth}
      \includegraphics[width=\textwidth]{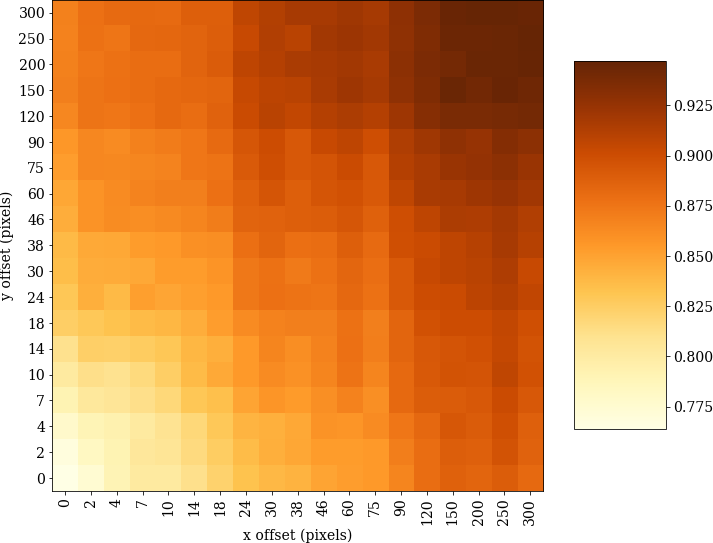}
      \caption{Prediction rate, $r_t$ \label{subfig:rt_08}}
    \end{subfigure}
    \hfill
    \begin{subfigure}[t]{0.45\textwidth}
      \includegraphics[width=\textwidth]{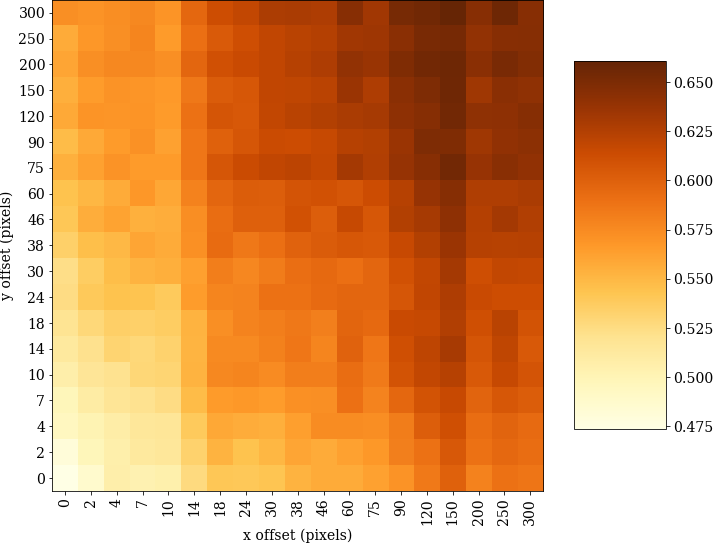}
      \caption{Accurately labeled prediction rate, $r_a$ \label{subfig:ra_08}}
    \end{subfigure}
    \begin{subfigure}[t]{0.45\textwidth}
      \includegraphics[width=\textwidth]{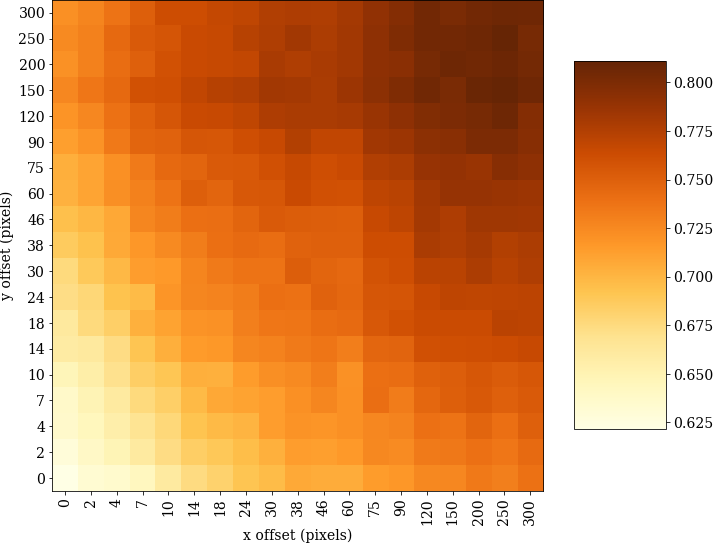}
      \caption{Average confidence score for the top prediction, $c_t$ \label{subfig:ct_08}}
    \end{subfigure}
    \hfill
    \begin{subfigure}[t]{0.45\textwidth}
      \includegraphics[width=\textwidth]{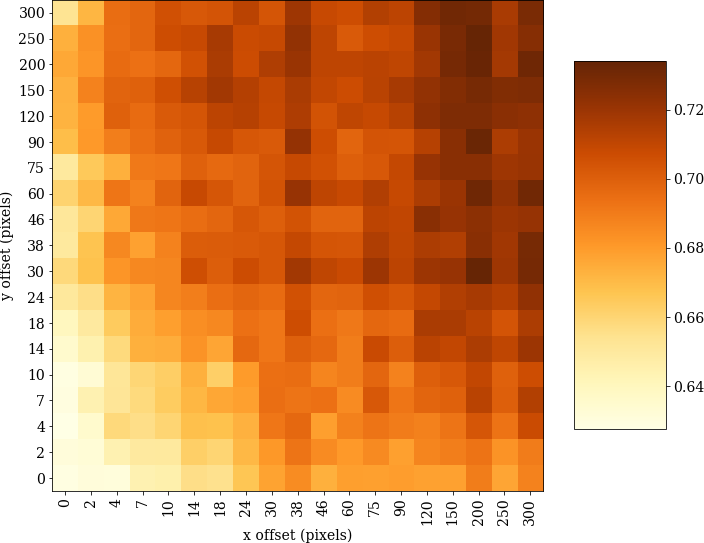}
      \caption{Average confidence score for the top accurately labeled prediction, $c_a$ \label{subfig:ca_08}}
    \end{subfigure}
    \begin{subfigure}[t]{0.45\textwidth}
      \includegraphics[width=\textwidth]{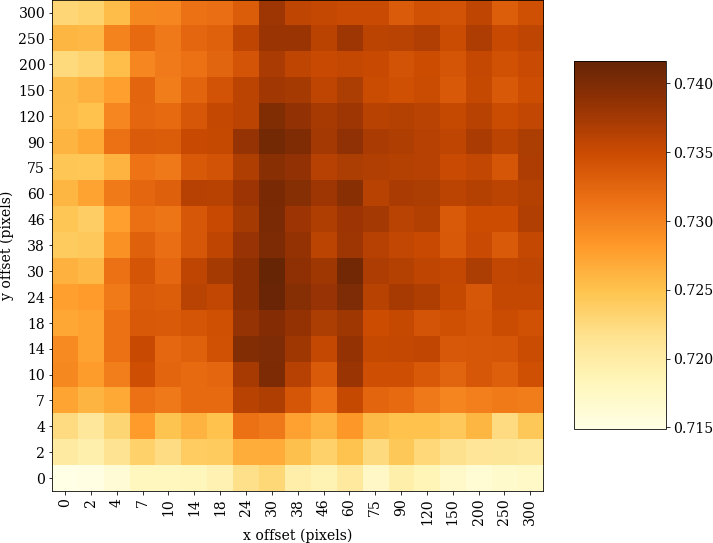}
      \caption{Average IoU score of the top prediction, $s_t$ \label{subfig:st_08}}
    \end{subfigure}
    \hfill
    \begin{subfigure}[t]{0.45\textwidth}
      \includegraphics[width=\textwidth]{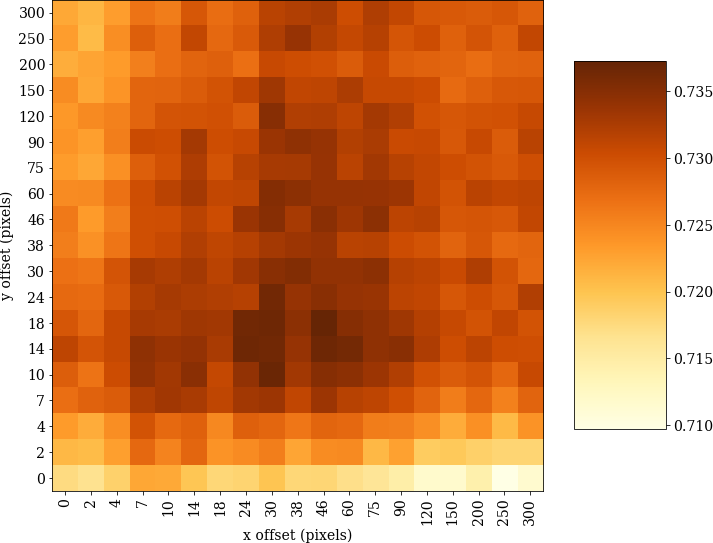}
      \caption{Average IoU score of an accurately labeled prediction, $s_a$ \label{subfig:sa_08}}
    \end{subfigure}
	\end{center}
  \caption{Mask R-CNN performance with respect to spatial position for objects with major dimension equal to $0.08$ times the width of the image crop ($64$ pixels).
	 \label{fig:res_8prop}}
\end{figure*}

As the object size increases the magnitude of this drop reduces, with the deficit along the non-corner boundary regions only amounting to a drop of $7\text{-}10$ percentage points on average for both $r_t$ and $r_a$ along the extreme periphery (Figures \ref{subfig:rt_08} and \ref{subfig:ra_08}, respectively), with performance beginning to improve $15\text{-}20$ pixels in from the boundary. Nevertheless, there still appears to be a consistent small drop in detection rate over the larger $75$ pixel band around the border seen on the order of $3$ percentage points for both $r_t$ and $r_a$.

Similar to the performance pattern seen for detection rates, average confidence scores assigned to object detections are on average lower near the image border and are worst in the corner. It is interesting to note that the drop in confidence relative to the image border seems to be more consistent and stronger for the top predictions, $c_t$, than for accurate predictions $c_a$. Nevertheless, this drop in confidence should be noted for applications which use confidence values for further reasoning about network output, particularly in the case of scenes with a potential mix of central and peripheral objects.

In contrast to confidence and detection rate, IoU scores do not follow a consistent pattern, and the degree of anisotropy across positions is relatively low. In fact, for the smallest objects the \emph{highest} average IoU scores are found in the corner and along the border in the $x$ direction in both the top prediction, $s_t$, and average prediction, $s_a$  conditions (Figures \ref{subfig:st_05} and \ref{subfig:sa_05}, respectively). This could be due to the fact that only the objects which are most obviously segmented from the background are detected in these positions, but this would require further investigation to confirm, particularly since the pattern switches for larger objects such that there is a very minor average deficit along the border, with the worst performance in the corners (the start of this shift is visible in Figures \ref{subfig:st_08} and \ref{subfig:sa_08}, and is clearer for the IoU scores of the larger objects presented in Section \ref{subsec:large_objects}). Nevertheless, these results may indicate that the RoIAlign component of Mask R-CNN \cite{HeEtAl2017}, which helps to project features back onto the pixel-level source for them in order to generate accurate segmentation maps, may help stabilize the performance of mask generation with respect to the image boundary when a successful detection takes place.

\subsection{Results for Large Objects}
\label{subsec:large_objects}

Large objects span many pixels, so even when a large object appears near an image boundary there will still be a number of pixels making up the object which are located well away from the boundary which can help overcome any degredation in peripheral representation and allow the network to successfully identify the object. Much as decreasing the average proportion of object occlusion decreases the performance drop caused by occlusion \cite{KoporecPers2019}, decreasing the proportion of the object falling within the portion of the image boundary affected by the network's peripheral anisotropy decreases the impact of this anisotropy. Thus, for tasks involving the whole object (such as the detection and segmentation task investigated in this work), it is perhaps not surprising that the magnitude of performance anisotropy exhibited gets smaller as the target object size increases.

\begin{figure*}[!htbp]
	\begin{center}
    \begin{subfigure}[t]{0.45\textwidth}
      \includegraphics[width=\textwidth]{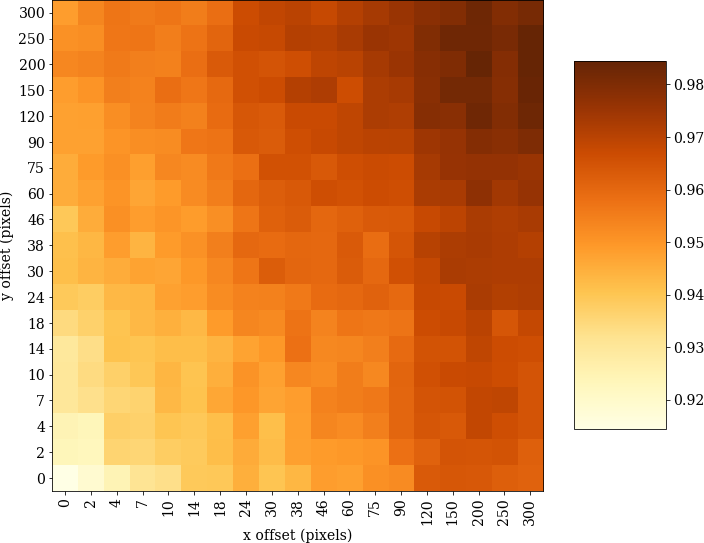}
      \caption{Prediction rate, $r_t$ \label{subfig:rt_12}}
    \end{subfigure}
    \hfill
    \begin{subfigure}[t]{0.45\textwidth}
      \includegraphics[width=\textwidth]{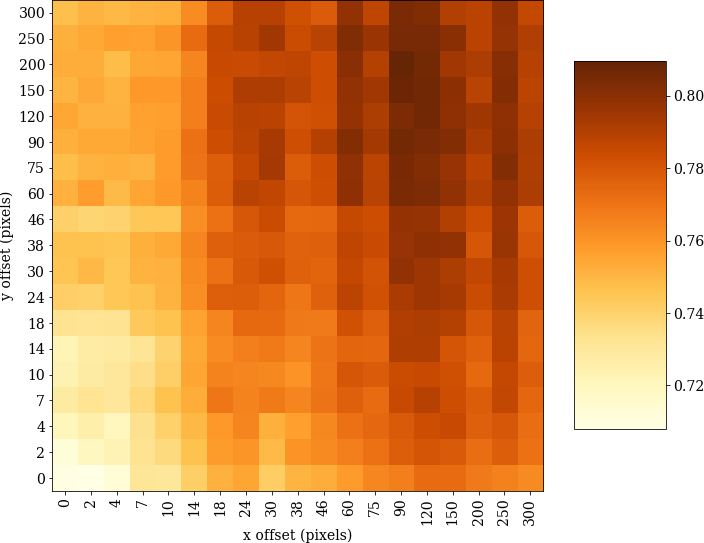}
      \caption{Accurately labeled prediction rate, $r_a$ \label{subfig:ra_12}}
    \end{subfigure}
    \begin{subfigure}[t]{0.45\textwidth}
      \includegraphics[width=\textwidth]{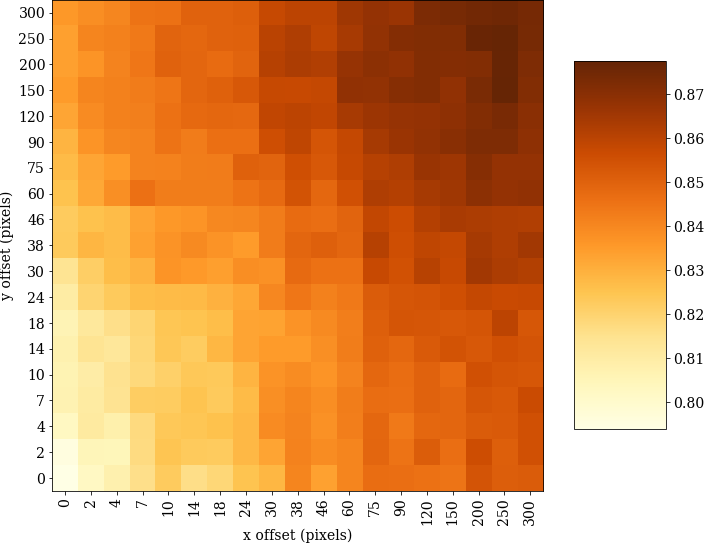}
      \caption{Average confidence score for the top prediction, $c_t$ \label{subfig:ct_12}}
    \end{subfigure}
    \hfill
    \begin{subfigure}[t]{0.45\textwidth}
      \includegraphics[width=\textwidth]{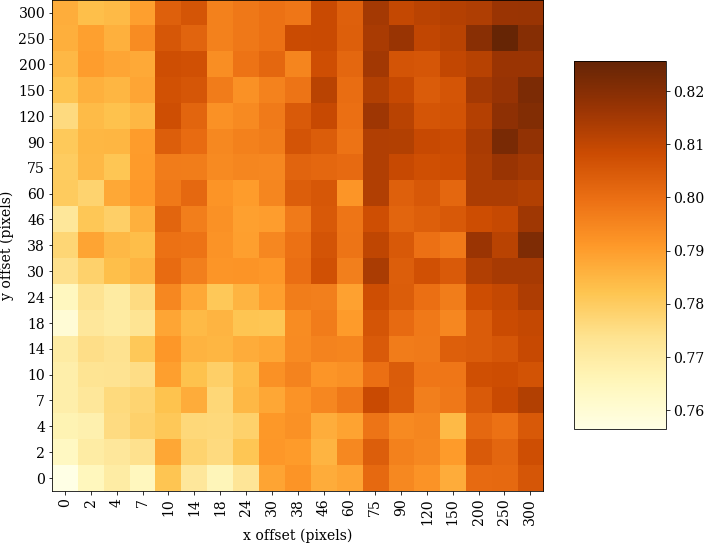}
      \caption{Average confidence score for the top accurately labeled prediction, $c_a$ \label{subfig:ca_12}}
    \end{subfigure}
    \begin{subfigure}[t]{0.45\textwidth}
      \includegraphics[width=\textwidth]{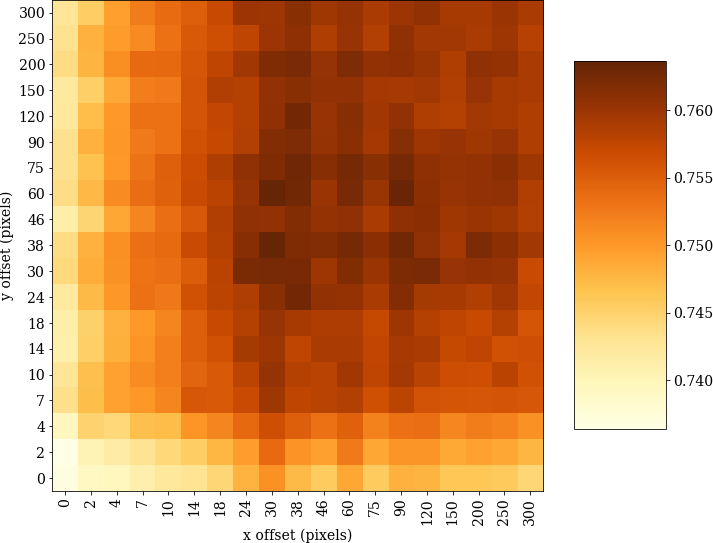}
      \caption{Average IoU score of the top prediction, $s_t$ \label{subfig:st_12}}
    \end{subfigure}
    \hfill
    \begin{subfigure}[t]{0.45\textwidth}
      \includegraphics[width=\textwidth]{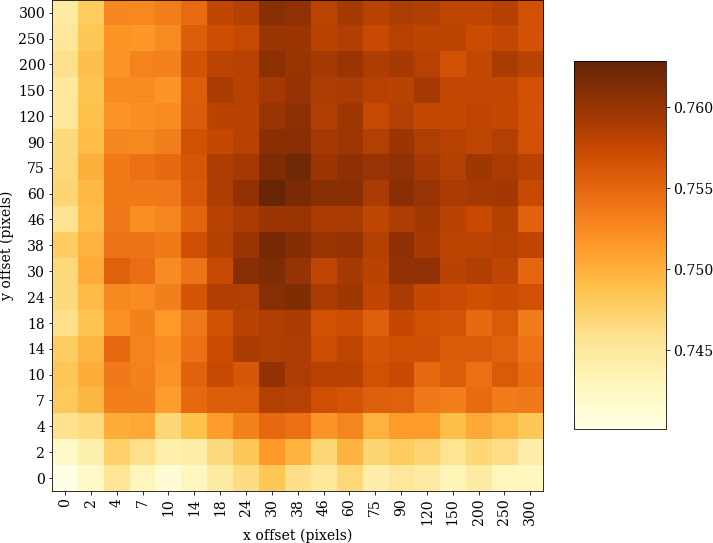}
      \caption{Average IoU score of an accurately labeled prediction, $s_a$ \label{subfig:sa_12}}
    \end{subfigure}
	\end{center}
  \caption{Mask R-CNN performance with respect to spatial position for objects with major dimension equal to $0.12$ times the width of the image crop ($96$ pixels).
	 \label{fig:res_12prop}}
\end{figure*}

Results for objects with major dimension equal to $0.12$ times the crop dimension ($96$ pixels) are shown in Figure \ref{fig:res_12prop}. The overall pattern of behavior for both the prediction rates and confidence scores is consistent with that found for smaller objects, but the magnitude of performance drop is only approximately $6$ percentage points in the valid prediction rate in the worst case (corner detections) and appears to only be a $2\text{-}3$ percentage point drop along the image boundary. The magnitude is slightly larger for accurate prediction rates, but still much reduced in comparison to smaller objects. Similarly, average confidence scores are affected as for smaller objects, but with a lower magnitude of performance drop. Interestingly, unlike what was seen for smaller objects, IoU scores appear to show a consistent deficit, albeit extremely small in magnitude, directly along the image periphery.

\begin{figure*}[!htbp]
	\begin{center}
    \begin{subfigure}[t]{0.45\textwidth}
      \includegraphics[width=\textwidth]{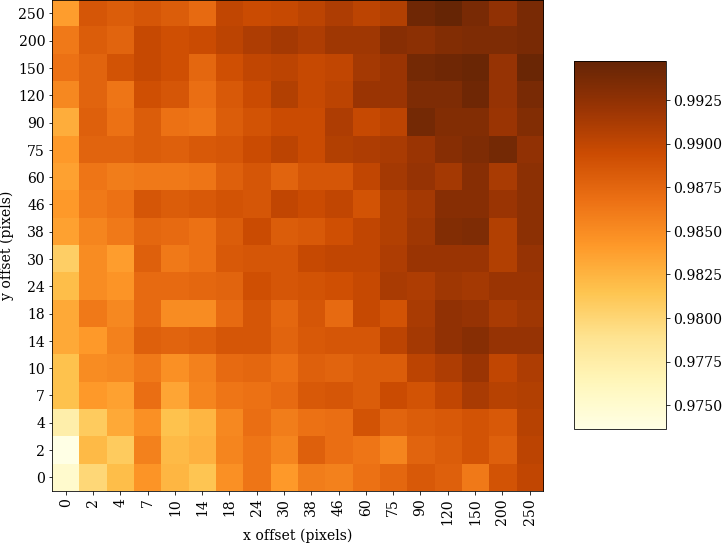}
      \caption{Prediction rate, $r_t$ \label{subfig:rt_18}}
    \end{subfigure}
    \hfill
    \begin{subfigure}[t]{0.45\textwidth}
      \includegraphics[width=\textwidth]{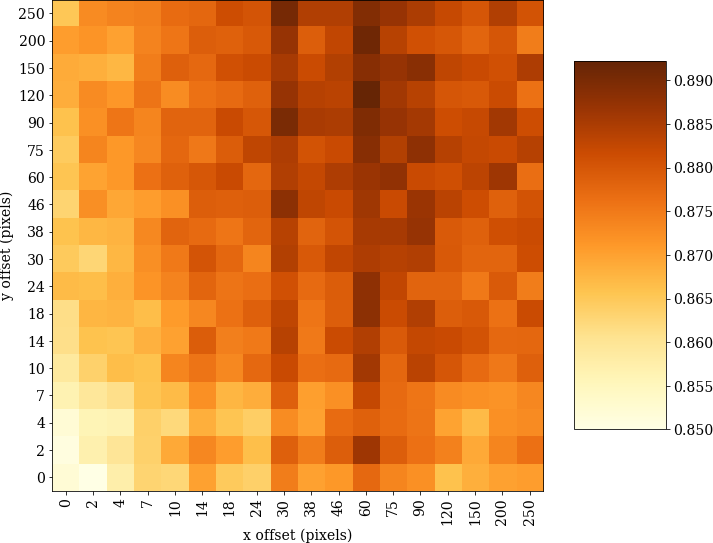}
      \caption{Accurately labeled prediction rate, $r_a$ \label{subfig:ra_18}}
    \end{subfigure}
    \begin{subfigure}[t]{0.45\textwidth}
      \includegraphics[width=\textwidth]{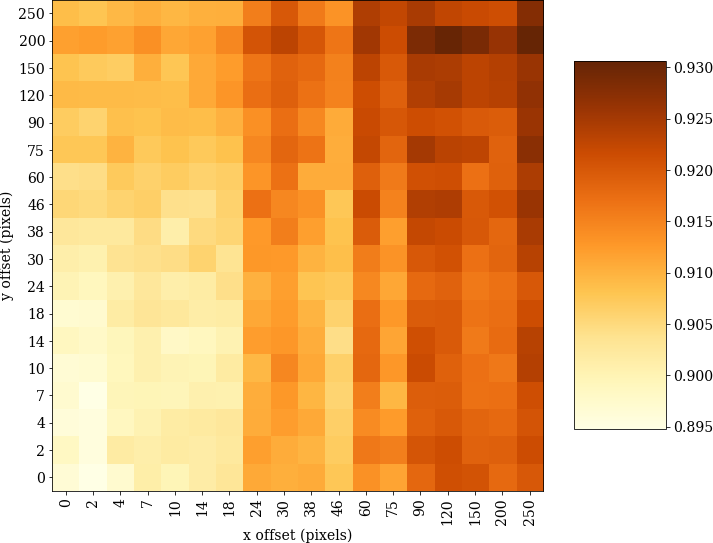}
      \caption{Average confidence score for the top prediction, $c_t$ \label{subfig:ct_18}}
    \end{subfigure}
    \hfill
    \begin{subfigure}[t]{0.45\textwidth}
      \includegraphics[width=\textwidth]{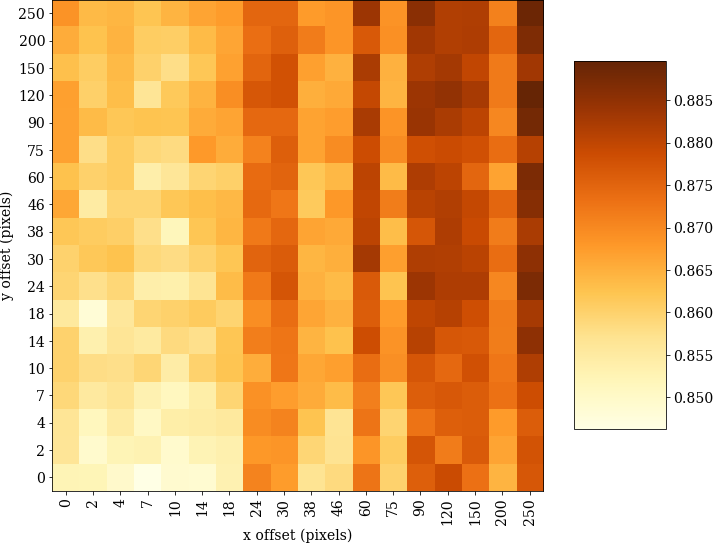}
      \caption{Average confidence score for the top accurately labeled prediction, $c_a$ \label{subfig:ca_18}}
    \end{subfigure}
    \begin{subfigure}[t]{0.45\textwidth}
      \includegraphics[width=\textwidth]{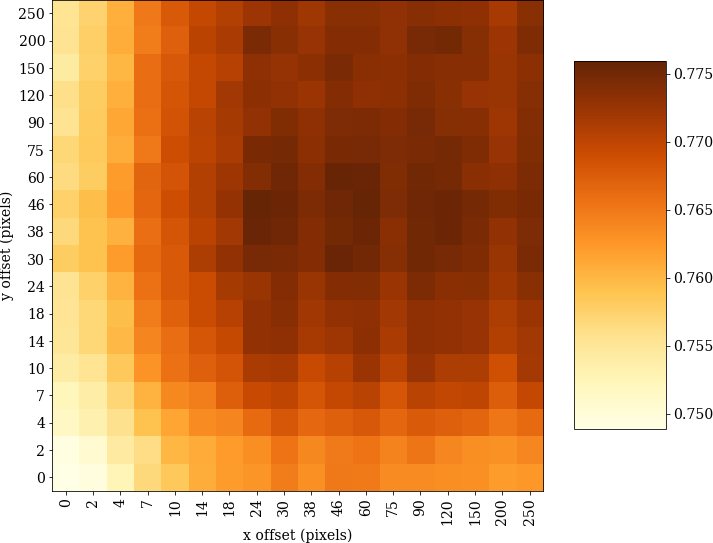}
      \caption{Average IoU score of the top prediction, $s_t$ \label{subfig:st_18}}
    \end{subfigure}
    \hfill
    \begin{subfigure}[t]{0.45\textwidth}
      \includegraphics[width=\textwidth]{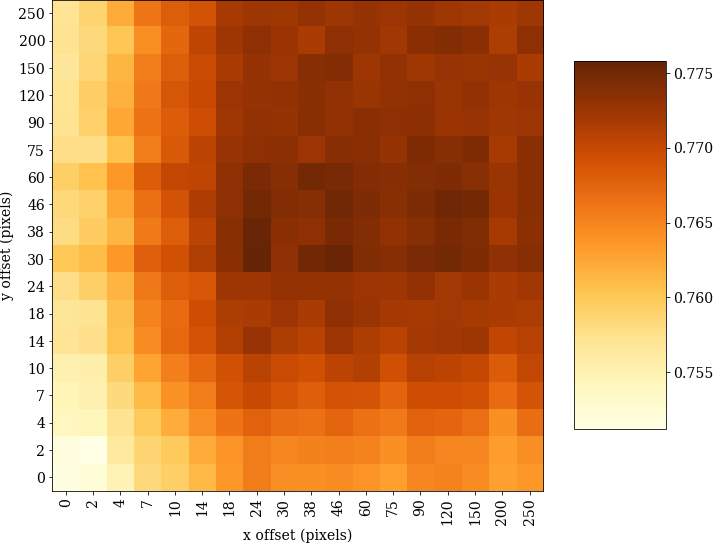}
      \caption{Average IoU score of an accurately labeled prediction, $s_a$ \label{subfig:sa_18}}
    \end{subfigure}
	\end{center}
  \caption{Mask R-CNN performance with respect to spatial position for objects with major dimension equal to $0.18$ times the width of the image crop ($144$ pixels).
	 \label{fig:res_18prop}}
\end{figure*}

For the largest size of objects tested (major dimension equal to $0.18$ times the crop dimension ($144$ pixels)), the peripheral deficits have almost fully disappeared (Figure \ref{fig:res_18prop}). Even in the worst case of objects located directly in the image corner, there is a less than $2$ percentage point drop in valid detection rate and a $3\text{-}5$ percentage point drop in accurate detection rates. Similarly, confidence scores show a small reduction in the corner and some indications of a slight reduction along the extreme periphery of the image, but the drop is marginal at most and far noisier than the behavioral pattern found for smaller objects. Oddly, the pattern in the IoU scores found for $96$ pixel objects is repeated, with a consistent but small reduction in IoU scores along the image boundary. It is unclear why this would appear far more consistently for larger target objects than for smaller ones, but the small magnitude suggests that this would only be a concern for applications which require extremely fine segmentation precision.

It is important to note that the detection and classification sub-tasks being performed by Mask R-CNN are essentially a binary decision task over the whole object; targets are detected and classified as whole units. Thus, as object size increases the peripheral performance deficit is reduced as the network can use the feature information from the more centrally located portion of a target to mitigate the representation of the peripheral components. However, for any task which relies either on a non-binary assessment of object appearance (such as quality assessment, \eg scoring the appeal of fruits and vegetables based on their color, texture, and lack of blemishes) or the extraction of internal detail (such as question answering, \eg finding a cat with a specific color of eyes), we expect that object size will no longer be sufficient to mitigate performance anisotropy.

\section{Discussion}

Despite the intent of convolutional networks to operate in a spatially invariant manner, image padding leads to measurable effects on network performance in the peripheral portions of the image. As network depth and complexity increases, the specific pattern of a network anisotropy becomes harder to analytically predict, and we have thus demonstrated a methodology for empirically characterizing and quantifying network behavior with respect to target size and spatial location.

There are a number of possible ways to incorporate analysis of this form into a vision system design. When camera positions relative to potential regions of interest are largely fixed, such as in out of stock detection \cite{ChaubardGarafulic2018} or person overboard monitoring \cite{KatsamenisEtAl2020}, understanding the spatial anisotropy of a network can allow the camera field of view to be placed in such a way so as to ensure that regions of interest are unlikely to fall outside of the network's highest performing regions. Similarly, when processing very high resolution images using a tiled approach (as in \cite{ChenEtAl2019}), characterizing network behavior allows tile overlap to be optimized in order to maximize performance while minimizing redundant processing.

For applications to more unconstrained environments, such as mobile robots, our results further support the need for active vision approaches to ensure that objects or regions of interest remain well localized within an agent's field of view when extracting information \cite{AndreopoulosTsotsos2013, BajcsyEtAl2017}. It is possible that a saliency-based gaze mechanism (such as \cite{WlokaEtAl2018}) could provide a general purpose interface between camera control and visual processing, or a strategy could be learned in order to accomplish specific task goals (\eg as in \cite{ChengEtAl2018}).

Finally, it would also be possible to use this methodology to evaluate different image padding techniques (such as zero-padding or reflection) for a given network and problem application. However, in order to do this the network would first have to be retrained independently with each padding style to ensure a fair comparison, and evaluation for spatial anisotropy as presented in this paper should also be done in conjunction with more classic benchmark evaluations given the findings of both Islam \etal \cite{IslamEtAl2020} and Kayhan \& Gemert \cite{KayhanGemert2020}, which demonstrate that networks are potentially using zero padding to encode absolute spatial location within an image. This may provide useful information for network predictions, and so it is possible that there is a tradeoff between peripheral performance and central performance with different styles of padding.

\subsubsection*{Acknowledgements}

The research in this paper was supported by the following grants held by John K. Tsotsos: the Air Force Office of Scientific Research (FA9550-18-1-0054), the Canada Research Chairs Program (950-231659), and the Natural Sciences and Engineering Research Council of Canada (RGPIN-2016-05352). Both authors are grateful for the support.

{\small
\bibliographystyle{ieee_fullname}
\bibliography{boundary_effect}
}

\end{document}